\documentclass[a4paper,twoside]{article}

\usepackage{epsfig}
\usepackage{subcaption}
\usepackage{calc}
\usepackage{amssymb}
\usepackage{amstext}
\usepackage{amsmath}
\usepackage{amsthm}
\usepackage{multicol}
\usepackage{pslatex}
\usepackage{apalike}
\usepackage{algorithm2e}
\usepackage[bottom]{footmisc}
\usepackage{booktabs}
\usepackage{doi}
\usepackage{SCITEPRESS}     

\begin{document}

\title{Video Summarization Techniques: A Comprehensive Review}

\author{\authorname{Toqa Alaa\sup{1}, Ahmad Mongy\sup{1}, Assem Bakr\sup{1}, Mariam Diab\sup{1}, and Walid Gomaa\sup{1,2}}
\affiliation{\sup{1}Department of Computer Science and Engineering, Egypt-Japan University of Science and Technology, Alexandria, Egypt}
\affiliation{\sup{2}Faculty of Engineering, Alexandria University, Alexandria, Egypt}
\email{\{toqa.alaa, ahmad.aboelnaga, asem.abdelhamid, mariam.diab, walid.gomaa\}@ejust.edu.eg}
}


\abstract{The rapid expansion of video content across a variety of industries, including social media, education, entertainment, and surveillance, has made video summarization an essential field of study. The current
work is a survey that explores the various approaches and methods created for video summarizing, emphasizing both abstractive and extractive strategies. The process of extractive summarization involves the identification of key frames or segments from the source video, utilizing methods such as shot boundary recognition, and clustering. On the other hand, abstractive summarization creates new content by getting the essential content from the video, using machine learning models like deep neural networks and natural language processing, reinforcement learning, attention mechanisms, generative adversarial networks, and multi-modal learning. We also include approaches that incorporate the two methodologies, along with discussing the uses and difficulties encountered in real-world implementations. The paper also covers the datasets used to benchmark these techniques. This review attempts to provide a state-of-the-art thorough knowledge of the current state and future directions of video summarization research. 
}

\onecolumn \maketitle \normalsize \setcounter{footnote}{0} \vfill

\section{\uppercase{Introduction}}
\label{sec:introduction}
\begin{figure}
    \centering
    \includegraphics[width=1\linewidth]{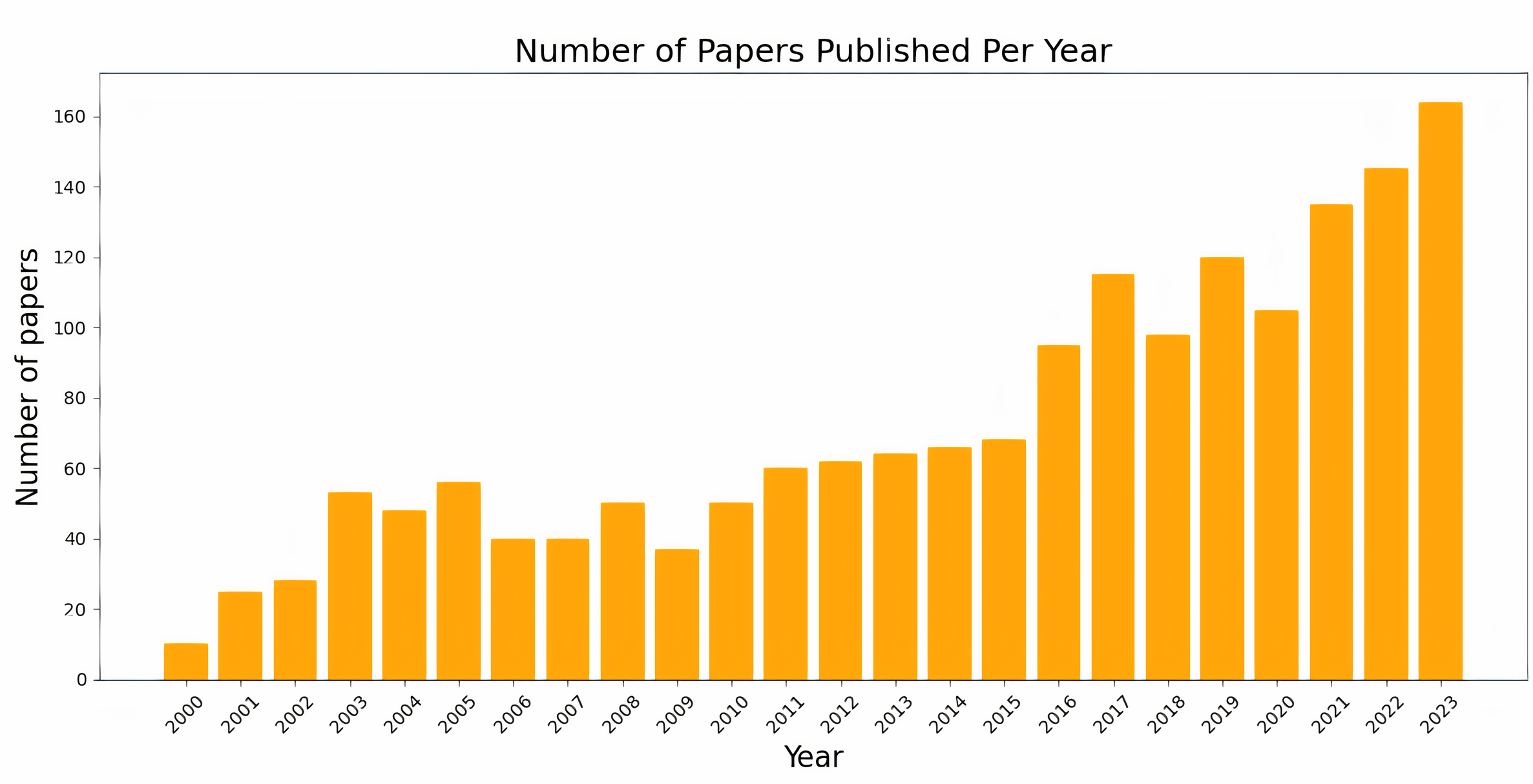}
    \caption{The number of papers per year containing in their title the phrase "video summarization" according to google scholar.}
    \label{fig: paper cnt}
\end{figure}

In recent years, technology has advanced rapidly, leading to camcorders being integrated into many devices \cite{pritch2008nonchronological}. People often capture their daily activities and special moments, creating large volumes of video content. With mobile devices making it easy to create and share videos on social media, there has been an explosion of videos available on the web \cite{rochan2019video}. Searching for specific video content and categorizing it can be very time-consuming. Traditional methods of representing a video as a series of consecutive frames work well for watching movies but have limitations for new multimedia services like content-based search, retrieval, navigation, and video browsing \cite{panagiotakis2009equivalent}. To address the need for efficient time management, automatic video content summarization and indexing techniques have been developed \cite{pritch2008nonchronological}. These techniques help with accessing, searching, categorizing, and recognizing actions in videos. The number of research papers on video summarization has been increasing yearly, as shown in Figure \ref{fig: paper cnt}.
According to the papers published, video summarization methods can be categorized into several key approaches. Static video summarization create a summary by selecting a collection of keyframes that correspond to significant scenes or events from the video. Static video summarization is employed in \cite{kumar2016equal} through detecting events by selecting keyframes using k-means clustering. Event-based summarization methods focuses on summarizing specific events within the video. These methods are useful when certain actions are the primary goal of the summary. The authors in \cite{banjar2024sports} focus on summarizing specific events in sports videos. In addition to static and event-based video summarization, there are personalized video summarization techniques that tailors the summary with respect to the preferences and the interests of the user. Topic-related summarizes are generated by \cite{zhu2023topic} to meet the subjective needs of different users.

Recent advancement in deep learning has further enhanced the capabilities of video summarization. For example, Attention mechanisms and reinforcement learning has been employed to generate concise and meaningful summaries. Attention mechanisms allow models to focus on relevant parts of the video. By dynamically focusing on different spatial and temporal features, these models can capture important models and details in the video. Reinforcement learning enables models to learn optimal summaries by interacting with the environment which is the video content. The RL agents learn to maximize the reward based on the quality of the generated summary \cite{zhou2018deep}.

Moreover, multi-modal learning has emerged as a promising direction in video summarization. By integrating information from multiple modalities such as audio, text, and visual features, multi-modal models can generate summaries that are richer in context \cite{zhu2023topic}. 

In addition to these advancements, researchers have explored single-view and multiView summarization strategies. Single-view summarization focuses on generating a concise representation from a single perspective of the video. In contrast, multiView summarization integrates information from multiple viewpoints \cite{parihar2021multiview}.

Lastly, the integration of interactive features in video summarization tools is becoming more frequent. Features like user feedback mechanisms allow users to customize and refine summaries based on their preferences and needs. This interactivity makes video summarization tools more versatile and user-friendly in diverse applications and contexts \cite{wu2022intentvizor}.

The paper is organized as follows. Section \ref{sec:introduction} provides an introduction, offering an overview of video summarization. Section \ref{sec:methodology} discusses methodologies employed in video summarization. Section \ref{sec:techniques} examines various techniques utilized in the field. Section \ref{sec:datasets} explores datasets utilized in the covered publications. Section \ref{sec:applications} explores applications of video summarization. Section \ref{sec:conclusion} summarizes our work and provides an outlook on future directions.

\section{\uppercase{Video Summarization Methodology}}
\label{sec:methodology}

Video summarization methodologies can be categorized into two main categories: extractive and abstractive methods.

\subsection{Extractive Methods}
Extractive video summarization methods select important segments from the original video without modifying the content.
\subsubsection{Keyframe Selection}
Extractive methods often involve selecting keyframes based on labeled data and criteria like objects, events, and user perception. keyframe selection can be branched into object-based, event-based, and perception-based keyframe selection. Object-based methods focuses on specific objects within the video \cite{meena2023review}. Event-based methods concentrate on particular moments and actions. Perception-based methods involve frames that are considered important according to the user perception \cite{kopru2022use}.
\subsubsection{Keyshot Selection}
Extractive summarization also involve keyshot selection, which divides the video into shots and selects representative shots. It can be divided into activity-based, object-based, event-based, and area-based keyshot selection. Activity-based shot selects shots that depict important activities or movements. Area-based shot selection focuses on specific regions within the video \cite{Mujtaba_2022}.

\subsection{Abstractive Methods}
Abstractive video summarization aims to create concise summaries of videos by generating new content that captures the essence of the original video. Unlike extractive methods, which select and combine together existing segments from the video, abstractive summarization involves interpreting and synthesizing information to create new, shorter representations \cite{he2023align}.

\section{\uppercase{Video summarization techniques}}
\label{sec:techniques}

This section discuss the various techniques used in video summarization. These techniques can be grouped according to the learning method, the type of the extracted features, the type of the input video, or the type of the output summary.
\subsection{Learning-based Methods}
Video summarization techniques can be classified according to the approach used. This section includes supervised, unsupervised, weakly supervised and reinforcement methods.
\subsubsection{Supervised Methods}
Supervised video summarization involves using annotated or labeled data during the training phase to learn how to generate summarizes. These methods involve datasets where each video is paired with a corresponding summary that highlights the most important events in the video. Supervised methods in video summarization typically uses Convolutional Neural Networks (CNNs), Recurrent Neural Networks (RNNs), Long Short-Term Memory (LSTM) Networks, or Transformer architecture ~\cite{he2023align,zhu2023topic,lin2022deep}.
\subsubsection{Unsupervised Methods}
Unsupervised video summarization attracts growing attention and has seen considerable progress through clustering methods, and generative adversarial networks. Unsupervised methods cover clustering-based techniques. The goal of clustering is to separate a finite unlabeled dataset into discrete and meaningful clusters.  Clustering was introduced to segment the video sequence in \cite{chen2023scene}. The key idea is to integrate Visual Place Recognition into the clustering process to enhance spatial diversity. Subsequently, a representative key-frame is selected from each cluster to serve as the summary.

\begin{figure}[!h]
  \vspace{-0.2cm}
  \centering
   \includegraphics[width=0.98\linewidth]{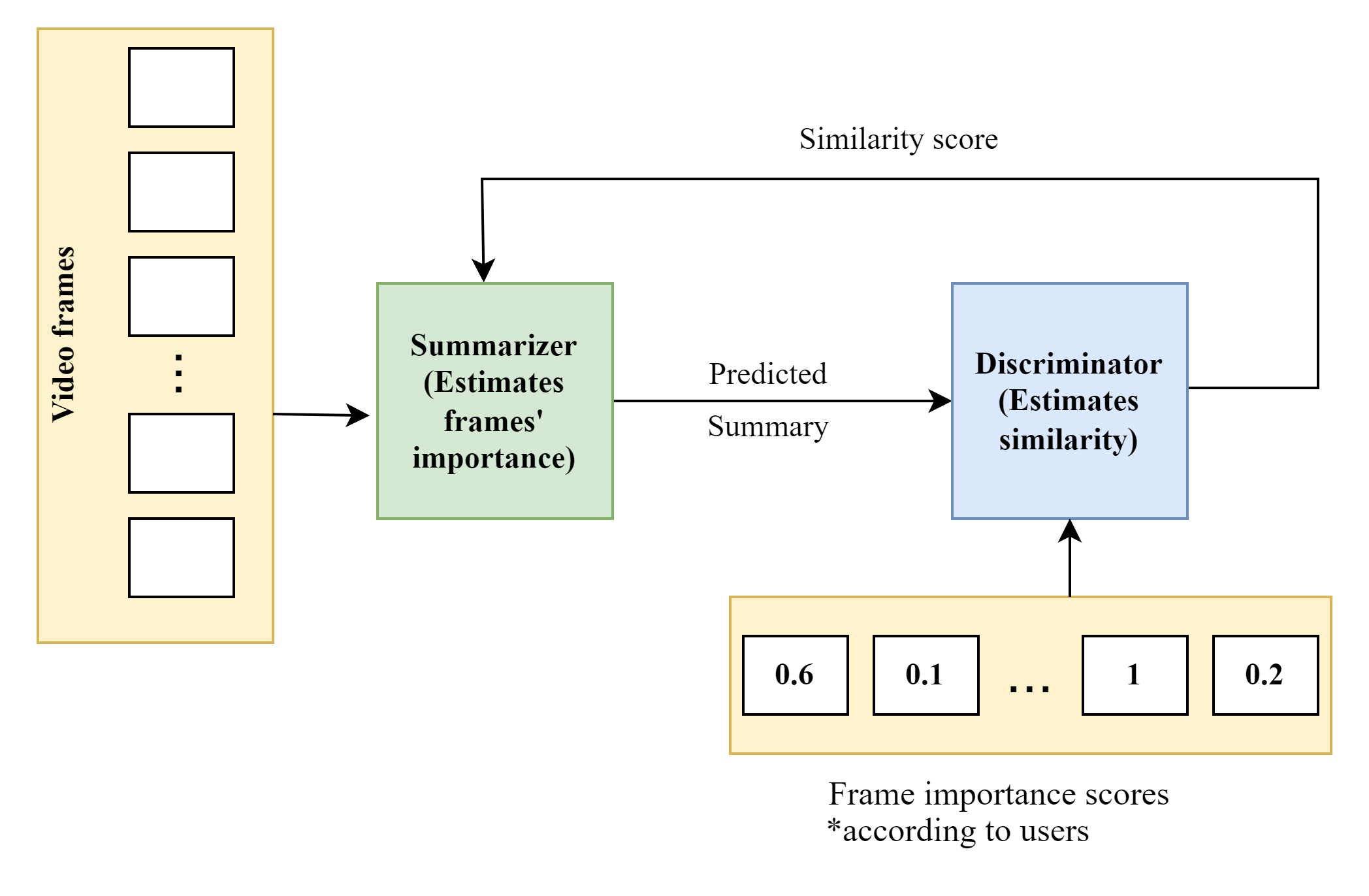}
  \caption{Generator and discriminator in video summarization process.}
  \label{fig:gans}
\end{figure}

Some techniques involve adversarial training through a generator and a discriminator. The Summarizer tries to fool the discriminator to not distinguish the predicted from the user generated summary, and the discriminator aims to learn how to make this distinction. When the discriminator’s confidence is very low, the Summarizer is able to generate a summary that is very close to the users’ expectations. Figure \ref{fig:gans} gives an overview of how the generator and the discriminator work to produce the video summary.
Fully Convolutional Sequence Network (FCSN) is used in selecting the key frames and a summary discriminator network is employed to differentiate between real and artificial
summaries through adversarial training in \cite{rochan2019video}.

Most of the methods that incorporate GANs to achieve Unsupervised LSTM units to learn how to assess the importance of each video frame. However, these methods resulted in findings about the low variation of the computed frame-level importance scores by LSTMs. As a consequence, the selections made by the trained LSTM seem to have a limited impact in summarization. The authors in \cite{he2019unsupervised} introduced a technique based on Attentive Conditional Generative Adversarial Networks (AC-GANs). This method uses a GAN framework where a BiLSTM predicts frame-level importance scores, and a discriminator differentiates these from raw frame features. A deep semantic and attentive network for Video Summarization (DSAVS) was proposed by ~\cite{lin2022deep} to generate unsupervised video summary by minimizing the distance between video representation and text representation, and introduce a self attention mechanism to capture the long-range temporal dependencies. The authors in \cite{apostolidis2020ac} proposed an architecture that embeds an Actor-Critic model into a Generative Adversarial Network and formulates the selection of important video fragments as a sequence generation task. The Actor and the Critic take part in a game that incrementally leads to the selection of the video key-fragments, and their choices at each step of the game result in a set of rewards from the discriminator. 

\subsubsection{Weakly Supervised Methods}
Video Summarization has a very high cost for data-labeling tasks, so it is desirable for the machine learning techniques to work with weak supervision. There are three types of weak supervision. The first is incomplete supervision, only a subset of training data is given with labels while the other data remain unlabeled. The second type is inexact supervision, i.e., only coarse-grained labels are given such as in~\cite{ye2021temporal} in which the model can learn to detect highlights by mining video characteristics with video level annotations (topic tags) only. The third type is inaccurate supervision, i.e., the given labels are not always ground-truth. Such a situation occurs, e.g., when the image annotator is careless or weary, or some images are difficult to categorize. A contextual temporal video encoder and a segment scoring transformer are used in \cite{narasimhan2022tl} to rank segments by their significance. This approach avoids the need for manual annotations and enhances scalability for large datasets.
\subsubsection{Reinforcement Learning Methods}
The process of video summarization is inherently sequential, as the choice of one segment can influence the importance and selection of subsequent segments. RL is well-suited for sequential decision-making tasks, enabling the model to optimize the selection process over time to achieve the best summary.

An end-to-end, reinforcement learning-based framework was proposed in \cite{zhou2018deep}. It designs a novel reward function that jointly accounts for diversity and representativeness of generated summaries and does not rely on labels or user interactions at all. However, this approach suffered from some limitations. The sparse reward problem leads to hard convergence as an agent receives the reward only after the whole summary is generated in conventional reinforcement learning methods. The authors in ~\cite{wang2024progressive} tried to address this problem by proposing a Progressive Reinforcement Learning network for Video Summarization (PRLVS) in an unsupervised learning fashion. In this method, the summarization task is decomposed hierarchically. The model trains an agent to modify the summary progressively. The agent chooses to replace one frame with some frame in the neighborhood and receives a reward for the whole summary at each step. 

Unsupervised video summarization with reinforcement learning and a 3D spatio-temporal U-Net is implemented by \cite{Liu2022} to efficiently encode spatio-temporal information of the input videos for downstream reinforcement learning.
\subsection{Input-Based Methods}
Video summarization techniques vary based on the type and modality of input videos. Inputs can differ in terms of the number of views—either singleView or multiView. Furthermore, variations in input modality also influence the approach to video summarization.
\subsubsection{SingleView Methods}
Single View video summarization has been a topic of great interest and active research for the last few decades. Many approaches have been proposed to generate effective summary. An equal partition-based clustering approach was proposed by \cite{kumur2016} to summarize single-view videos. It detects the events by selecting key-frames through K-means clustering algorithm. A key-frame based single-view video summarization method is proposed by \cite{tirupathamma2017key}. Frame difference is calculated between adjacent frames to discard redundant frames. Frames with difference greater than a certain threshold are considered as key-frames. An abnormality-type-based approach proposed by \cite{lin2015summarizing} targets surveillance videos by extracting the abnormal events. A patch-based method first models the normal patterns and the key regions of the events. Abnormal objects are detected in a scene by blob sequence optimization by considering spatial, temporal, size, and motion correlations among objects.
\subsubsection{MultiView Methods}
 MultiView video summarization is crucial in abstracting essential information from multiple videos of the same location and time. Single-view summarizers ignore the temporal order by processing simultaneously recorded views in a sequential order to fit as a single-view input. This results in redundant and repetitive summaries that do not exhibit the multi-stream nature of the footage. According to~\cite{elfeki2022multi}, there are two types of important events to be included in the universal summary. First, events where multiple views have a substantial overlap, in which the summary includes the most representative view. Second, events that are spatially independent, in which each view is processed separately from the rest. The authors in \cite{gong2014diverse} proposed the sequential determinantal point process (seqDPP), a probabilistic model for diverse sequential subset selection. The model seqDPP heeds the inherent sequential structures in video data, thus overcoming the deficiency of the standard DPP, which treats video frames as randomly permutable items. Meanwhile, seqDPP retains the power of modeling diverse subset. A novel view to video summarization is suggested by \cite{chu2015video} that exploits visual co-occurrence across multiple videos by finding shots that co-occur most frequently across videos retrieved using a topic keyword. The task of video summarization is defined in \cite{meng2017video} as a multiview representative selection problem. The main objective is to find visual elements that represent the video across different views. Optimization is done by the Multiview Sparse Dictionary Selection with Centroid Co-regularization (MSDS-CC) method. Another pioneering method is devised by~\cite{pan2018video} for getting a summary of custom length. Using a bottom-up algorithm known as video clip growth, it generates video abstract by an accurate frame adding process, allowing the users to customize the length of the video summaries.  A deep multiview video summarization network with an attention mechanism was proposed by \cite{ji2019video} to address the supervised video summarization problem. The encoding mechanism uses a Bidirectional Long Short-term Memory (Bi-LSTM) to encode the contextual information of the video frames, and two attention-based LSTM networks are used as decoders. The authors in ~\cite{parihar2021multiview} introduces another new approach for shot boundary detection using the frame similarity measures Jaccard and Dice. Using the Balanced Iterative Reducing and Clustering using Hierarchies (BIRCH) algorithms to reduce the redundant frames. The authors in ~\cite{elfeki2022multi} present a robust framework that identifies a diverse set of important events among moving cameras that often are not capturing the same scene, and selects the most representative view(s) at each event to be included in a universal summary by proposing a new adaptation of the widely used Determinantal Point Process \cite{gong2014diverse}. Multi-DPP, generalizes it to accommodate multi-stream setting while maintaining the temporal order.
\subsubsection{Modality-Based Methods}
For complex scenes in videos, it is difficult for a single modality to provide sufficient information, which may lead to incorrect prediction results. Synthesising the knowledge of multiple modalities makes up for the deficiency of single modality, and thus benefits making further decisions and improving the quality of summaries. This approach benefits from the rich, complementary information provided by each modality, enhancing the effectiveness and accuracy of the summarization process. Videos can have different types of modalities such as visual, audio, and textual modalities.

\begin{itemize}
\item Visual Modality: identifying different scenes and their transitions helps in segmenting the video into meaningful parts, each of which can be summarized separately. Additionally, detecting objects, people, and activities in the video provides valuable insights into the content and context of scenes, highlighting visually important moments. 

\item Audio Modality: utilizing automatic speech recognition (ASR) converts spoken words into text, facilitating the identification of crucial dialogues, keywords, and phrases. This process emphasizes significant discussions or narrative elements within the video.

\item Textual Modality: involving the representation of spoken content through textual means, including subtitles and other textual overlays accompanying the video, provides additional context and enhances understanding of the video's content.
\end{itemize}

Recently, multi-modal learning has proved that audio and vision modalities share a consistency space, and there are semantic relations between them. The audio modality can assist the vision modality to better understand the video content and structure. Concretely, the audio and vision are complementary to present activities in different modalities. The authors in \cite{zhao2021audiovisual} tries to encode the visual and the audio features using LSTM. The two-stream LSTM is utilized in the first stage to encode the audio and visual features sequentially, and capture their temporal dependency. Then, the audiovisual fusion LSTM is developed to exploit the consistency space between audio and visual information, and fuse them with an adaptive gating mechanism. A low-rank audio-visual tensor fusion mechanism was developed by \cite{ye2021temporal} to capture the complex association between the two modalities, which can efficiently generate informative audio-visual fused features. This audio-visual tensor overcomes the challenges encountered in linear fusion methods and in bilinear pooling. Linear fusion was not able to fully capture the complex association between the two modalities due to the distinct feature distribution of each modality since audio does not always correspond to the visual frames. A new technique was introduce by \cite{ye2021temporal}  called the AudioVisual Recurrent Network (AVRN) to improve video summarization by using both audio and visual information. Unlike traditional methods that focus only on visual data, AVRN combines audio and visual signals to provide a better understanding of video content and structure. 
AVRN was tested on the SumMe and TVsum datasets, and the results showed that combining audio and visual features improved video summarization compared to using only visual features.

The authors in \cite{yuan2017video} extract semantic information from the side, including video titles, user query, video description and user comments, and define a video summary by maximizing the relevance between visual and semantic features in a common latent space. The authors in \cite{wei2018video} use manual description annotations for videos and select video shots by minimizing the distance between the generated description sentence of the summary and the human annotated text of the original video, with the help of semantic attended networks. Both visual and textual data are leveraged in \cite{palaskar2019multimodal} to generate comprehensive summaries by extracting visual features from videos using pre-trained action recognition model, which captures key elements and actions. Simultaneously, textual data is obtained through transcriptions, which can be either user-generated or output from the automatic speech recognition (ASR) system using a pre-trained ASR model. Alanguage-guided multimodal transformer was proposed by \cite{narasimhan2021clip} that learns to score frames in a video based on their importance relative to one another and their correlation with a user-defined query or an automatically generated dense video caption. An alignment-guided self-attention module was proposed by \cite{he2023align} to align the temporal correspondence between video and text modalities and fuse cross-modal information in a unified manner. And it formulates a two novel contrastive losses to model both inter-sample and intra-sample correlations to model the cross-modal correlation at different granularities.

The authors in \cite{ghauri2020classification} introduced computational models that predict the importance of segments in (lengthy) videos. It uses Bi-LSTM layers to incorporate information from each modality. A novel multimodal Transformer model was introduces by \cite{zhu2023topic} which can adaptively fuse multimodal features to make up for the deficiency of single modality, thus benefits making further decisions and improving the quality of summaries. The method consists of a feature extraction module, a feature learning module, a topic classification module and a frame selection module.
\subsection{Features-Based Methods}
This section divides the video summarization techniques according to the type of the features included in the used approach.
\subsubsection{Spatial-Features-Based Methods}
Using spatial features for video summarization involves analyzing and extracting information based on the spatial properties of the video frames. Spatial features capture the visual content within each frame, providing valuable data that can be used to identify important scenes and key frames.

The approach in \cite{Mahmoud2013} relies on clustering color and texture features extracted from the video frames using a modified Density-based clustering non-parametric algorithm 
(DBSCAN) algorithm~\cite{Ester1996ADA} to summarize the video content. The original video undergoes pre-sampling. Next, color features are extracted using a color histogram in the HSV color space, and texture features are extracted using a two-dimensional Haar wavelet transform in HSV color space. Video frames are then clustered with a modified DBSCAN algorithm, and key frames are selected. Finally, these key frames are arranged in their original order to aid visual understanding.
\subsubsection{Temporal-Features-Based Summarization} Temporal-based Summarization involves dividing the video into temporal segments or shots. This can be achieved through various methods such as scene change detection, where abrupt changes in frames indicate the beginning or end of a segment. 
The authors in \cite{lee2012} used clustering of frame color histograms to segment temporal events. 
Temporal video segmentation is used in \cite{Potapov2014} for detecting shot or scene boundaries. This approach takes into account the differences between all pairs of frames. 
A novel hierarchical temporal context encoder was introduced in \cite{ye2021temporal} to embed local temporal clues in between neighboring segments.
\subsubsection{Spatio-temporal-Features-Based Summarization}
Spatio-temporal summarization involves capturing and representing spatial and temporal information from videos to create a concise summary. It has a very critical role in the management of the vast amounts of video data generated in various fields, allowing the extraction of essential content without manually sifting through all the footage.  A new technique is introduced in \cite{Li2021} called SUM-GDA (SUMmarization via Global Diverse Attention) that builds a convolutional neural network with an attention mechanism to generate informative and globally diverse video summaries. The authors in \cite{lin2022deep} introduced a new framework for video summarization called Deep Hierarchical LSTM Networks with Attention (DHAVS). This methodology extracts spatio-temporal features using 3D ResNeXt-101 and employs a deep hierarchical LSTM network with an attention mechanism to capture long-range dependencies and focus on important keyframes. Additionally, it proposes a cost-sensitive loss function to improve the selection of critical frames under class imbalance. DHAVS was tested on the SumMe and TVSum datasets, and the results show that DHAVS outperforms state-of-the-art methods, demonstrating significant advancements in video summarization.
\begin{figure*}
    \centering
    \includegraphics[width=1\linewidth]{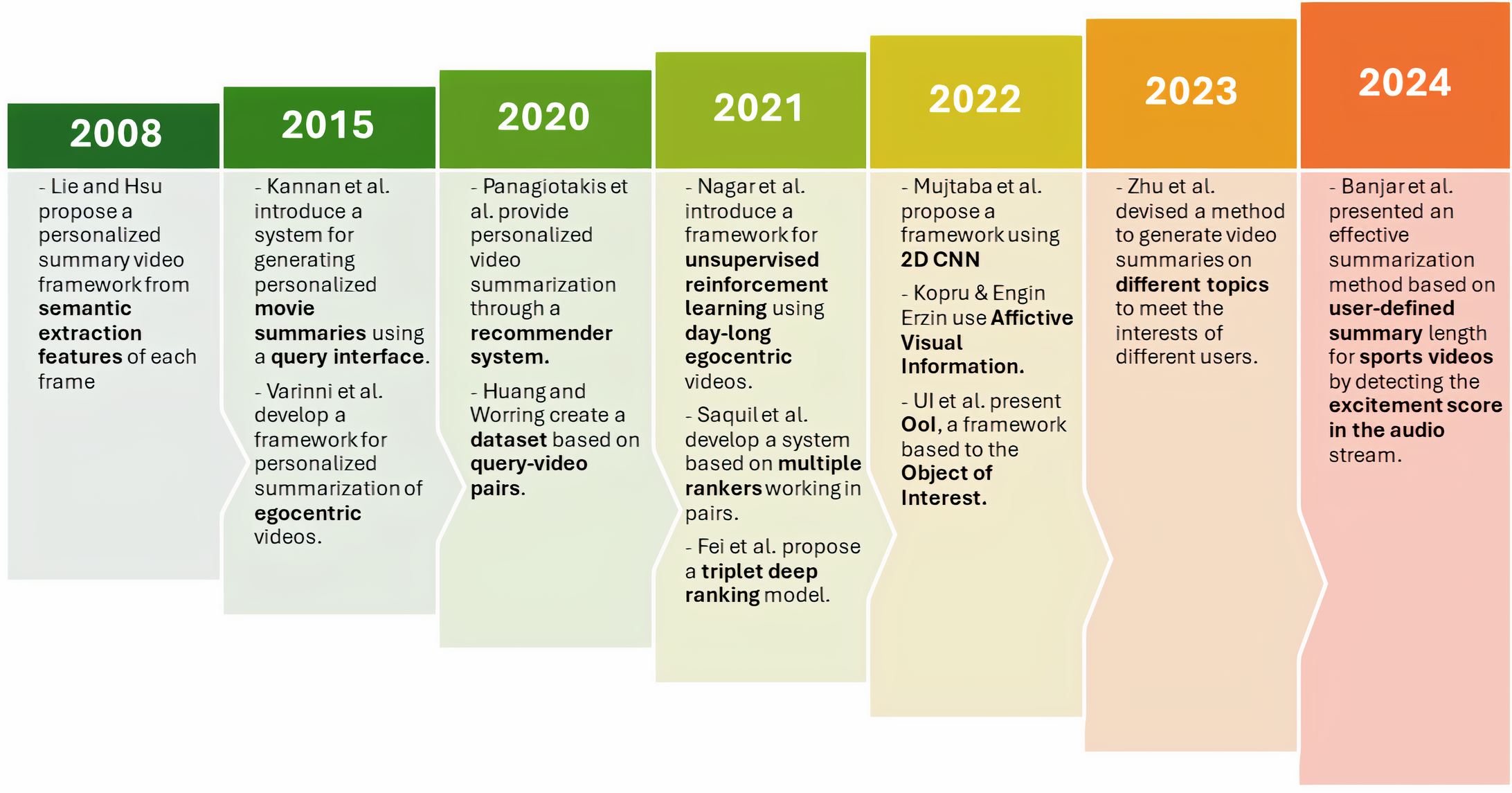}
    \caption{Milestones in personalized video summarization.}
    \label{fig:milestones}
\end{figure*}
\subsection{Output-Based Methods}
This section discusses the video summarization methods according to the type of the output summary.
\subsubsection{Generic Video Summarization}
A generic summary may be obtained by applying either storyboard summarization method or video skimming method. In both of these methods, a generic summary is created by picking up the parts of the video that are considered to depict the content of the video in a synopsis form. This type of summary is useful when the user does not have an intuition about the content of the video and just wishes to quickly go through highlights of the video. What makes a part of a video qualifying to be in the summary is a debatable topic in generic video summarization. Simple generic summary may be created by sampling the frames either uniformly or randomly. But an advanced approach of selecting the frames which represent the semantics of the video well can be developed using video processing and computer vision techniques. 

The authors in ~\cite{basavarajaiah2021gvsum} proposed a generic video summarization technique, called GVSUM which can be applied to a variety of videos like surveillance videos, movies, sports videos, home videos etc. First, it extracts the component frames of the video. Instead of decoding the entire video, I-frames (which are a type of frame in video encoding that contains complete information about a single frame of video) from the encoded video are extracted using partial decoding of the video. Visual features from these frames are extracted and based on these visual features, all the frames are assigned a unique cluster number using unsupervised
grouping of the frames. Whenever there is a change in cluster number of the frame, that frame is selected to be a part of the summary. The change in cluster number represents the change in visual scene of the video as the frames are grouped based on their visual features.

\subsubsection{Personalized Video Summarization}
Personalized video summarization is an emerging technology designed to create customized short versions of videos tailored to an individual viewer's preferences, interests, and needs. 
The authors in~\cite{zhu2023topic} focus on generating multiple topic-related video summaries, which are especially advantageous in meeting the subjective needs.  The topic classification module aims to classify the updated visual features into topic classes. Since one video frame may present multiple topics, the topic classification is actually a multi-label classification task. Accordingly, the topic classification module is decomposed into multiple binary classifiers, where each binary classifier predicts whether the input frame belongs to the corresponding topic. Figure \ref{fig:milestones} presents the milestones in personalized video summarization. It also indicates the first published paper on personalized video summaries and the long gap before many related works started being published. 

Personalized video summarization is useful for creating sports highlights so that each viewer can view the content that interests her for each sport activity. The authors in \cite{tejero2018summarization} proposed a framework that uses neural networks to generate a personalized video summarization that can be applied to any sport in which a game has a sequence of actions. To select the highlights of the original video, they use the actions of the players as a contract.  An effective summarization method is presented in \cite{banjar2024sports} based on user-defined summary length for sports videos by detecting the excitement score in the audio stream.

\subsubsection{Input-Query-Based Video Summarization}
Query-based video summarization is an approach that aims to generate concise and informative summaries of videos based on specific user queries or topics of interest. Unlike traditional video summarization methods that may rely on predefined rules or algorithms to extract key frames or segments, query-based summarization takes into account user-defined criteria to tailor the summary to the user's needs.

The authors in \cite{Xiao2020} formulate the task as a problem of computing similarity between video shots and query, and propose a convolutional network with local
self-attention mechanism and query-aware global attention mechanism to learn visual information of each shot. A new framework ``IntentVizor'' is introduces in \cite{wu2022intentvizor} that is designed for interactive video summarization, integrating both textual and visual queries. It consists of two modules: the intent module, which extracts the user intent, and the summary module, which performs the summarization. Both modules utilize a shared backbone called Granularity-Scalable Ego-Graph Convolutional Network (GSE-GCN), effectively aligning video features with generic queries or intents. This framework supports multi-modal queries, enabling users to interactively adjust the summarization process by representing intent as an adjustable distribution over learned basis intents rather than fixed categories. 

\section{\uppercase{Datasets}}
\label{sec:datasets}

In video summarization research, the choice of datasets is crucial for training and evaluating the performance of various techniques. Table~\ref{table:evaluation_metrics} outlines the  key datasets commonly used in the video summarization techniques, highlighting their characteristics, contents, and relevance to different summarization methods.

\begin{table*}[htbp]
\centering
\caption{Datasets for video summarization.}
\label{table:evaluation_metrics}
\begin{tabular}{p{2cm} p{5cm} p{2cm} p{2cm} p{2cm}}
\toprule
\textbf{Dataset} & \textbf{Description} & \textbf{Size} & \textbf{Duration (min)} & \textbf{Modality}  \\
\midrule
TVSum 
\cite{song2015tvsum}
& Title-based containing news, how-to, documentary, vlog, egocentric genres and 1,000 annotations of shot-level importance scores obtained via crowdsourcing. & 50  & 2-10 & Uni \\
\midrule
SumMe 
\cite{gygli2014creating}
& Each video is annotated with at least 15 human summaries.  & 25 & 1-6  & Uni \\
\midrule
BLiSS 
\cite{he2023align}
& Contains livestream videos and transcripts with multimodal summaries (674 videos, 12,629 text)  & 13,303 & 5  & Multi \\
\midrule

TopicSum 
\cite{zhu2023topic}
& Contains frame-level importance scores, and topic labels annotations.  & 136 & 5  & Multi \\
\midrule
Multi-Ego 
\cite{elfeki2022multi}
& Contains videos recorded simultaneously by three cameras, covering a wide variety of real-life scenarios. & 41 & 3-7  & Uni \\
\midrule
How2
\cite{palaskar2019multimodal}
& Short instructional videos from different domains, each video is accompanied by a transcript. & 80,000 & 1-2 & Multi\\
\midrule
EDUVSUM 
\cite{ghauri2020classification}
& Educational videos with subtitles from three popular e-learning platforms: Edx,YouTube, and TIB AV-Portal. & 98 & 10-60 & Multi \\
\midrule
COIN
\cite{tang2019coin}
& Contains 11,000 instructional videos covering 180 tasks. Used for training by creating pseudo summaries. & 8,521 & 3 & Uni \\
\midrule
FineGym 
\cite{shao2020finegym}
& A hierarchical video dataset for fine-grained action understanding. & 156 & 10 & Multi \\
\midrule
CrossTask
\cite{xu2019crosstask}
& Contains 4,700 instructional videos covering 83 tasks; used for generating pseudo summaries and training. & 3,675 & 3 & Uni \\
\midrule
WikiHow Summaries
\cite{hassan2018wikihow}
& High-quality test set created by scraping WikiHow articles that include video demonstrations and visual depictions of steps. & 2,106 & 2-7 & Multi \\
\midrule
HowTo100M 
\cite{miech2019howto100m}
& A large-scale dataset with more than 100 million video clips from YouTube, annotated with textual descriptions. & 136M & 6-7 & Multi \\
\midrule
VTW 
\cite{zeng2016title}
& Large dataset containing annotated videos from YouTube with highlighted key shots & 2,529 & 2-5 & Uni \\
\bottomrule \\

\end{tabular} 
\end{table*}
\section{\uppercase{Applications}}
\label{sec:applications}

Video summarization has a wide range of applications across various industries, revolutionizing how video content is managed and utilized. This section explores the different kinds of applications of video summarization.
\subsection{Video Highlight Detection}

Streaming platforms use summarization techniques to create trailers and highlight reels, helping viewers get a quick overview of content. This can entice potential viewers to watch full episodes or movies. The rise of short-form video sharing applications has attracted world-wide attention on the Internet. An automated method is needed desperately to identify highlight clips from untrimmed  long videos~\cite{ye2021temporal}.

\subsection{Personalized Recommendations}
By analyzing the topics and themes present in video summaries, recommendation systems can tailor suggestions to individual users' viewing habits. For instance, if a user frequently watches summaries of cooking videos, the platform can prioritize recommending new cooking videos or related content, thereby providing a more personalized experience. The authors in \cite{zhu2023topic} devised a method to generate video summaries on different topics to meet the interests of different users. Topic-aware video summarization takes into account the subjectivity of video summarization and allows users to have more choices,
rather than generating a single summary of an input video in existing methods. 

\subsection{Security and Monitoring}

Video summarization is a crucial technology in the fields of security and monitoring, where it significantly enhances the efficiency and effectiveness of surveillance systems. The authors in \cite{wu2022intentvizor} proposed IntentVizor which improves the interactivity of video summarization that aids in the process of monitoring.

\subsection{Video Indexing}
Video indexing is a crucial process in the field of multimedia management that involves organizing and tagging video content to facilitate efficient search, retrieval, and analysis. It plays a significant role in various applications, from media and entertainment to security and education. Summarized videos provide a condensed version of the content, making it easier for search algorithms to process and index key themes and topics \cite{Tiwari2021}.
Summarization can facilitate the extraction of relevant keywords and phrases that are crucial for indexing, enabling more precise search results \cite{Apostolidis2021}. Summarized videos require less storage space and bandwidth for indexing systems, which can lead to cost savings and faster processing times. 

\subsection{Anomaly Detection}
Leading video anomaly detection methods often rely on large-scale training datasets with long training times. By leveraging pre-trained deep models and denoising autoencoders (DAEs), video summarization enables faster model deployment while maintaining comparable detection performance \cite{Sultani_2018_CVPR}.
The proposed framework in \cite{wu2022intentvizor} explains each anomaly detection result in surveillance videos, providing insights into the decision-making process

\subsection{Education and E-Learning} 
Video summarization can facilitate learning by providing summarizes of lengthy lectures, tutorials, or educational videos. Students can quickly grasp key concepts, review essential points, and focus on areas of interest, enhancing the overall learning experience \cite{Benedetto2023}.

\subsection{Social Media and Marketing} 

Social media platforms can utilize video summarization to create promotional content, or product demonstrations. These summaries help capture viewers' attention and send key messages effectively in a short time \cite{Sabha2023}.

\section{\uppercase{Conclusions}}
\label{sec:conclusion}

Video summarization is a crucial area of research as it contributes to many valuable applications such as video retrieval, personalized video recommendation, and in surveillance systems. 

Video summarization techniques can be categorized according to a lot of aspects. The techniques used vary according to the type of the input video, or the type of the output summary. This paper tried to provide a comprehensive review on existing summarization techniques, methodologies, and benchmark datasets. 

Despite the numerous papers and various techniques developed for video summarization, several challenges remain. There is a need for new datasets that support multi-modal input and output for different topics. For example, there is a need to build datasets to support creating textual and visual summarization for educational, medical, and news broadcasts videos. In addition, the area of interactive video summarization has to be developed further. The user must has the freedom to control different parameters of the output summary such as the length of the produced summary. To address this challenge, future research should focus on developing advanced frameworks that allow for more user interaction and customization in the video summarization process. Moreover, developing real-time techniques could be highly beneficial for practical applications such as autonomous driving and security surveillance. Furthermore, there is a need to develop scalable systems that can handle the large volumes of video data.
What's more, addressing the challenges of context-awareness and semantic understanding in video summarization can lead to more meaningful and accurate summaries. This involves developing models that can understand the narrative structure, detect key events, and identify salient information across different modalities.

In conclusion, while significant progress has been made in the field of video summarization, future work must focus on enhancing user interaction, developing multi-modal datasets, improving real-time capabilities, and advancing semantic understanding to fully realize the potential of video summarization in various applications.



\bibliographystyle{apalike}
{\small
\bibliography{example}}

\begin{thebibliography}{}

\bibitem[Apostolidis et~al., 2020]{apostolidis2020ac}
Apostolidis, E., Adamantidou, E., Metsai, A.~I., Mezaris, V., and Patras, I. (2020).
\newblock Ac-sum-gan: Connecting actor-critic and generative adversarial networks for unsupervised video summarization.
\newblock {\em IEEE Transactions on Circuits and Systems for Video Technology}, 31(8):3278--3292.

\bibitem[Apostolidis et~al., 2021]{Apostolidis2021}
Apostolidis, E., Adamantidou, E., Metsai, A.~I., Mezaris, V., and Patras, I. (2021).
\newblock Video summarization using deep neural networks: A survey.
\newblock {\em Proceedings of the IEEE}, 109(11):1838–1863.

\bibitem[Banjar et~al., 2024]{banjar2024sports}
Banjar, A., Dawood, H., Javed, A., and Zeb, B. (2024).
\newblock Sports video summarization using acoustic symmetric ternary codes and svm.
\newblock {\em Applied Acoustics}, 216:109795.

\bibitem[Basavarajaiah and Sharma, 2021]{basavarajaiah2021gvsum}
Basavarajaiah, M. and Sharma, P. (2021).
\newblock Gvsum: generic video summarization using deep visual features.
\newblock {\em Multimedia Tools and Applications}, 80(9):14459--14476.

\bibitem[Benedetto et~al., 2023]{Benedetto2023}
Benedetto, I., La~Quatra, M., Cagliero, L., Canale, L., and Farinetti, L. (2023).
\newblock Abstractive video lecture summarization: applications and future prospects.
\newblock {\em Education and Information Technologies}, 29(3):2951–2971.

\bibitem[Chen et~al., 2023]{chen2023scene}
Chen, C., Zhu, M., Singh, A.~P., Yan, Y., Xu, F.~J., and Feng, C. (2023).
\newblock Scene summarization: Clustering scene videos into spatially diverse frames.
\newblock {\em arXiv preprint arXiv:2311.17940}.

\bibitem[Chu et~al., 2015]{chu2015video}
Chu, W.-S., Song, Y., and Jaimes, A. (2015).
\newblock Video co-summarization: Video summarization by visual co-occurrence.
\newblock In {\em Proceedings of the IEEE conference on computer vision and pattern recognition}, pages 3584--3592.

\bibitem[Elfeki et~al., 2022]{elfeki2022multi}
Elfeki, M., Wang, L., and Borji, A. (2022).
\newblock Multi-stream dynamic video summarization.
\newblock In {\em Proceedings of the IEEE/CVF Winter Conference on Applications of Computer Vision}, pages 339--349.

\bibitem[Ester et~al., 1996]{Ester1996ADA}
Ester, M., Kriegel, H.-P., Sander, J., and Xu, X. (1996).
\newblock A density-based algorithm for discovering clusters in large spatial databases with noise.
\newblock In {\em Knowledge Discovery and Data Mining}.

\bibitem[Ghauri et~al., 2020]{ghauri2020classification}
Ghauri, J.~A., Hakimov, S., and Ewerth, R. (2020).
\newblock Classification of important segments in educational videos using multimodal features.
\newblock {\em arXiv preprint arXiv:2010.13626}.

\bibitem[Gong et~al., 2014]{gong2014diverse}
Gong, B., Chao, W.-L., Grauman, K., and Sha, F. (2014).
\newblock Diverse sequential subset selection for supervised video summarization.
\newblock {\em Advances in neural information processing systems}, 27.

\bibitem[Gygli et~al., 2014]{gygli2014creating}
Gygli, M., Grabner, H., Riemenschneider, H., and Van~Gool, L. (2014).
\newblock Creating summaries from user videos.
\newblock In {\em Computer Vision--ECCV 2014: 13th European Conference, Zurich, Switzerland, September 6-12, 2014, Proceedings, Part VII 13}, pages 505--520. Springer.

\bibitem[Hassan et~al., 2018]{hassan2018wikihow}
Hassan, S., Saleh, M., Kubba, M., et~al. (2018).
\newblock Wikihow: A large scale text summarization dataset.
\newblock In {\em Proceedings of the 2018 Conference on Empirical Methods in Natural Language Processing}, pages 3243--3252.

\bibitem[He et~al., 2023]{he2023align}
He, B., Wang, J., Qiu, J., Bui, T., Shrivastava, A., and Wang, Z. (2023).
\newblock Align and attend: Multimodal summarization with dual contrastive losses.
\newblock In {\em Proceedings of the IEEE/CVF Conference on Computer Vision and Pattern Recognition}, pages 14867--14878.

\bibitem[He et~al., 2019]{he2019unsupervised}
He, X., Hua, Y., Song, T., Zhang, Z., Xue, Z., Ma, R., Robertson, N., and Guan, H. (2019).
\newblock Unsupervised video summarization with attentive conditional generative adversarial networks.
\newblock In {\em Proceedings of the 27th ACM International Conference on multimedia}, pages 2296--2304.

\bibitem[Ji et~al., 2019]{ji2019video}
Ji, Z., Xiong, K., Pang, Y., and Li, X. (2019).
\newblock Video summarization with attention-based encoder--decoder networks.
\newblock {\em IEEE Transactions on Circuits and Systems for Video Technology}, 30(6):1709--1717.

\bibitem[K{\"o}pr{\"u} and Erzin, 2022]{kopru2022use}
K{\"o}pr{\"u}, B. and Erzin, E. (2022).
\newblock Use of affective visual information for summarization of human-centric videos.
\newblock {\em IEEE Transactions on Affective Computing}.

\bibitem[Kumar et~al., 2016a]{kumar2016equal}
Kumar, K., Shrimankar, D.~D., and Singh, N. (2016a).
\newblock Equal partition based clustering approach for event summarization in videos.
\newblock In {\em 2016 12th International conference on signal-image technology \& internet-based systems (SITIS)}, pages 119--126. IEEE.

\bibitem[Kumar et~al., 2016b]{kumur2016}
Kumar, K., Shrimankar, D.~D., and Singh, N. (2016b).
\newblock Equal partition based clustering approach for event summarization in videos.
\newblock In {\em 2016 12th International Conference on Signal-Image Technology \& Internet-Based Systems (SITIS)}, pages 119--126.

\bibitem[Lee et~al., 2012]{lee2012}
Lee, Y.~J., Ghosh, J., and Grauman, K. (2012).
\newblock Discovering important people and objects for egocentric video summarization.
\newblock In {\em 2012 IEEE Conference on Computer Vision and Pattern Recognition}, pages 1346--1353.

\bibitem[Li et~al., 2021]{Li2021}
Li, P., Ye, Q., Zhang, L., Yuan, L., Xu, X., and Shao, L. (2021).
\newblock Exploring global diverse attention via pairwise temporal relation for video summarization.
\newblock {\em Pattern Recognition}, 111:107677.

\bibitem[Lin et~al., 2022]{lin2022deep}
Lin, J., Zhong, S.-h., and Fares, A. (2022).
\newblock Deep hierarchical lstm networks with attention for video summarization.
\newblock {\em Computers \& Electrical Engineering}, 97:107618.

\bibitem[Lin et~al., 2015]{lin2015summarizing}
Lin, W., Zhang, Y., Lu, J., Zhou, B., Wang, J., and Zhou, Y. (2015).
\newblock Summarizing surveillance videos with local-patch-learning-based abnormality detection, blob sequence optimization, and type-based synopsis.
\newblock {\em Neurocomputing}, 155:84--98.

\bibitem[Liu et~al., 2022]{Liu2022}
Liu, T., Meng, Q., Huang, J.-J., Vlontzos, A., Rueckert, D., and Kainz, B. (2022).
\newblock Video summarization through reinforcement learning with a 3d spatio-temporal u-net.
\newblock {\em IEEE Transactions on Image Processing}, 31.

\bibitem[Mahmoud et~al., 2013]{Mahmoud2013}
Mahmoud, K.~M., Ismail, M.~A., and Ghanem, N.~M. (2013).
\newblock {\em VSCAN: An Enhanced Video Summarization Using Density-Based Spatial Clustering}, page 733–742.
\newblock Springer Berlin Heidelberg.

\bibitem[Meena et~al., 2023]{meena2023review}
Meena, P., Kumar, H., and Yadav, S.~K. (2023).
\newblock A review on video summarization techniques.
\newblock {\em Engineering Applications of Artificial Intelligence}, 118:105667.

\bibitem[Meng et~al., 2017]{meng2017video}
Meng, J., Wang, S., Wang, H., Yuan, J., and Tan, Y.-P. (2017).
\newblock Video summarization via multi-view representative selection.
\newblock In {\em Proceedings of the IEEE international conference on computer vision workshops}, pages 1189--1198.

\bibitem[Miech et~al., 2019]{miech2019howto100m}
Miech, A., Laptev, I., and Sivic, J. (2019).
\newblock Howto100m: Learning a text-video embedding by watching hundred million video clips.
\newblock In {\em Proceedings of the IEEE/CVF International Conference on Computer Vision}, pages 2630--2640.

\bibitem[Mujtaba et~al., 2022]{Mujtaba_2022}
Mujtaba, G., Malik, A., and Ryu, E.-S. (2022).
\newblock Ltc-sum: Lightweight client-driven personalized video summarization framework using 2d cnn.
\newblock {\em IEEE Access}, 10:103041–103055.

\bibitem[Narasimhan et~al., 2022]{narasimhan2022tl}
Narasimhan, M., Nagrani, A., Sun, C., Rubinstein, M., Darrell, T., Rohrbach, A., and Schmid, C. (2022).
\newblock Tl; dw? summarizing instructional videos with task relevance and cross-modal saliency.
\newblock In {\em European Conference on Computer Vision}, pages 540--557. Springer.

\bibitem[Narasimhan et~al., 2021]{narasimhan2021clip}
Narasimhan, M., Rohrbach, A., and Darrell, T. (2021).
\newblock Clip-it! language-guided video summarization.
\newblock {\em Advances in neural information processing systems}, 34:13988--14000.

\bibitem[Palaskar et~al., 2019]{palaskar2019multimodal}
Palaskar, S., Libovick{\`y}, J., Gella, S., and Metze, F. (2019).
\newblock Multimodal abstractive summarization for how2 videos.
\newblock {\em arXiv preprint arXiv:1906.07901}.

\bibitem[Pan et~al., 2018]{pan2018video}
Pan, G., Qu, X., Lv, L., Guo, S., and Sun, D. (2018).
\newblock Video clip growth: A general algorithm for multi-view video summarization.
\newblock In {\em Advances in Multimedia Information Processing--PCM 2018: 19th Pacific-Rim Conference on Multimedia, Hefei, China, September 21-22, 2018, Proceedings, Part III 19}, pages 112--122. Springer.

\bibitem[Panagiotakis et~al., 2009]{panagiotakis2009equivalent}
Panagiotakis, C., Doulamis, A., and Tziritas, G. (2009).
\newblock Equivalent key frames selection based on iso-content principles.
\newblock {\em IEEE Transactions on circuits and systems for video technology}, 19(3):447--451.

\bibitem[Parihar et~al., 2021]{parihar2021multiview}
Parihar, A.~S., Pal, J., and Sharma, I. (2021).
\newblock Multiview video summarization using video partitioning and clustering.
\newblock {\em Journal of Visual Communication and Image Representation}, 74:102991.

\bibitem[Potapov et~al., 2014]{Potapov2014}
Potapov, D., Douze, M., Harchaoui, Z., and Schmid, C. (2014).
\newblock {\em Category-Specific Video Summarization}, page 540–555.
\newblock Springer International Publishing.

\bibitem[Pritch et~al., 2008]{pritch2008nonchronological}
Pritch, Y., Rav-Acha, A., and Peleg, S. (2008).
\newblock Nonchronological video synopsis and indexing.
\newblock {\em IEEE transactions on pattern analysis and machine intelligence}, 30(11):1971--1984.

\bibitem[Rochan and Wang, 2019]{rochan2019video}
Rochan, M. and Wang, Y. (2019).
\newblock Video summarization by learning from unpaired data.
\newblock In {\em Proceedings of the IEEE/CVF conference on computer vision and pattern recognition}, pages 7902--7911.

\bibitem[Sabha and Selwal, 2023]{Sabha2023}
Sabha, A. and Selwal, A. (2023).
\newblock Data-driven enabled approaches for criteria-based video summarization: a comprehensive survey, taxonomy, and future directions.
\newblock {\em Multimedia Tools and Applications}, 82(21):32635–32709.

\bibitem[Shao et~al., 2020]{shao2020finegym}
Shao, D., Zhao, Y., Dai, B., and Lin, D. (2020).
\newblock Finegym: A hierarchical video dataset for fine-grained action understanding.
\newblock In {\em Proceedings of the IEEE/CVF conference on computer vision and pattern recognition}, pages 2616--2625.

\bibitem[Song et~al., 2015]{song2015tvsum}
Song, Y., Vallmitjana, J., Stent, A., and Jaimes, A. (2015).
\newblock Tvsum: Summarizing web videos using titles.
\newblock In {\em Proceedings of the IEEE conference on computer vision and pattern recognition}, pages 5179--5187.

\bibitem[Sultani et~al., 2018]{Sultani_2018_CVPR}
Sultani, W., Chen, C., and Shah, M. (2018).
\newblock Real-world anomaly detection in surveillance videos.
\newblock In {\em Proceedings of the IEEE Conference on Computer Vision and Pattern Recognition (CVPR)}.

\bibitem[Tang et~al., 2019]{tang2019coin}
Tang, Z., Xiong, Y., Xu, Y., Wang, W., Hua, X.-S., and Zhang, J. (2019).
\newblock Coin: A large-scale dataset for comprehensive instructional video analysis.
\newblock In {\em Proceedings of the IEEE/CVF Conference on Computer Vision and Pattern Recognition}, pages 1207--1216.

\bibitem[Tejero-de Pablos et~al., 2018]{tejero2018summarization}
Tejero-de Pablos, A., Nakashima, Y., Sato, T., Yokoya, N., Linna, M., and Rahtu, E. (2018).
\newblock Summarization of user-generated sports video by using deep action recognition features.
\newblock {\em IEEE Transactions on Multimedia}, 20(8):2000--2011.

\bibitem[Tirupathamma, 2017]{tirupathamma2017key}
Tirupathamma, S. (2017).
\newblock Key frame based video summarization using frame difference.
\newblock {\em International Journal of Innovative Computer Science \& Engineering}, 4(3).

\bibitem[Tiwari and Bhatnagar, 2021]{Tiwari2021}
Tiwari, V. and Bhatnagar, C. (2021).
\newblock A survey of recent work on video summarization: approaches and techniques.
\newblock {\em Multimedia Tools and Applications}, 80(18):27187–27221.

\bibitem[Wang et~al., 2024]{wang2024progressive}
Wang, G., Wu, X., and Yan, J. (2024).
\newblock Progressive reinforcement learning for video summarization.
\newblock {\em Information Sciences}, 655:119888.

\bibitem[Wei et~al., 2018]{wei2018video}
Wei, H., Ni, B., Yan, Y., Yu, H., Yang, X., and Yao, C. (2018).
\newblock Video summarization via semantic attended networks.
\newblock In {\em Proceedings of the AAAI conference on artificial intelligence}, volume~32.

\bibitem[Wu et~al., 2022]{wu2022intentvizor}
Wu, G., Lin, J., and Silva, C.~T. (2022).
\newblock Intentvizor: Towards generic query guided interactive video summarization.
\newblock In {\em Proceedings of the IEEE/CVF Conference on Computer Vision and Pattern Recognition}, pages 10503--10512.

\bibitem[Xiao et~al., 2020]{Xiao2020}
Xiao, S., Zhao, Z., Zhang, Z., Yan, X., and Yang, M. (2020).
\newblock Convolutional hierarchical attention network for query-focused video summarization.
\newblock {\em Proceedings of the AAAI Conference on Artificial Intelligence}, 34(07).

\bibitem[Xu et~al., 2019]{xu2019crosstask}
Xu, Z., Buch, S., Ramanan, D., and Niebles, J.~C. (2019).
\newblock Cross-task weakly supervised learning from instructional videos.
\newblock In {\em Proceedings of the IEEE/CVF Conference on Computer Vision and Pattern Recognition}, pages 3537--3545.

\bibitem[Ye et~al., 2021]{ye2021temporal}
Ye, Q., Shen, X., Gao, Y., Wang, Z., Bi, Q., Li, P., and Yang, G. (2021).
\newblock Temporal cue guided video highlight detection with low-rank audio-visual fusion.
\newblock In {\em Proceedings of the IEEE/CVF International Conference on Computer Vision}, pages 7950--7959.

\bibitem[Yuan et~al., 2017]{yuan2017video}
Yuan, Y., Mei, T., Cui, P., and Zhu, W. (2017).
\newblock Video summarization by learning deep side semantic embedding.
\newblock {\em IEEE Transactions on Circuits and Systems for Video Technology}, 29(1):226--237.

\bibitem[Zeng et~al., 2016]{zeng2016title}
Zeng, K.-H., Chen, T.-H., Niebles, J.~C., and Sun, M. (2016).
\newblock Title generation for user generated videos.
\newblock In {\em Computer Vision--ECCV 2016: 14th European Conference, Amsterdam, The Netherlands, October 11-14, 2016, Proceedings, Part II 14}, pages 609--625. Springer.

\bibitem[Zhao et~al., 2021]{zhao2021audiovisual}
Zhao, B., Gong, M., and Li, X. (2021).
\newblock Audiovisual video summarization.
\newblock {\em IEEE Transactions on Neural Networks and Learning Systems}, 34(8):5181--5188.

\bibitem[Zhou et~al., 2018]{zhou2018deep}
Zhou, K., Qiao, Y., and Xiang, T. (2018).
\newblock Deep reinforcement learning for unsupervised video summarization with diversity-representativeness reward.
\newblock In {\em Proceedings of the AAAI conference on artificial intelligence}, volume~32.

\bibitem[Zhu et~al., 2023]{zhu2023topic}
Zhu, Y., Zhao, W., Hua, R., and Wu, X. (2023).
\newblock Topic-aware video summarization using multimodal transformer.
\newblock {\em Pattern Recognition}, 140:109578.

\end{thebibliography}



\end{document}